\documentclass[conference,transmag]{IEEEtran}
\IEEEoverridecommandlockouts
\usepackage{subcaption}
\usepackage{cite}
\usepackage{amsmath,amssymb,amsfonts}
\usepackage{algorithmic}
\usepackage{graphicx}
\usepackage{textcomp}
\usepackage{xcolor}
\usepackage{url}
\usepackage{booktabs} 
\usepackage[justification=centering]{caption} 
\usepackage{eso-pic}
\usepackage{lipsum}
\def\BibTeX{{\rm B\kern-.05em{\sc i\kern-.025em b}\kern-.08em
    T\kern-.1667em\lower.7ex\hbox{E}\kern-.125emX}}
\begin{document}

\AddToShipoutPictureBG*{%
  \AtPageUpperLeft{%
    \hspace{\paperwidth}%
    \raisebox{-\baselineskip}{%
      \makebox[0pt][r]{\tiny{\textcopyright 2022 IEEE. Personal use of this material is permitted. Permission from IEEE must be obtained for all other uses, in any current or future media, including reprinting/republishing this material for advertising or promotional purposes, creating new collective works, for resale or redistribution }}
}}}%
\AddToShipoutPictureBG*{%
  \AtPageUpperLeft{%
    \hspace{\paperwidth}%
    \raisebox{-1.5\baselineskip}{%
      \makebox[0pt][r]{\tiny{to servers or lists, or reuse of any copyrighted component of this work in other works.} }
}}}%

\title{DarkSLAM: GAN-assisted Visual SLAM  \\ for Reliable Operation in Low-light Conditions\\
}

\author{\IEEEauthorblockN{Alena Savinykh,
Mikhail Kurenkov, 
Evgeny Kruzhkov,  
Evgeny Yudin, \\
Andrei Potapov, 
Pavel Karpyshev, and 
Dzmitry Tsetserukou}
\IEEEauthorblockA{\textit{ISR Laboratory, Skolkovo Institute of Science and Technology, Moscow, Russia}}
\IEEEauthorblockA{$\{$Alena.Savinykh, Mikhail.Kurenkov, Evgeny.Kruzhkov, Evgeny.Yudin, \\
Andrei.Potapov, Pavel.Karpyshev, D.Tsetserukou$\}$@skoltech.ru}}

\maketitle

\begin{abstract}
\emph{Abstract} — Existing visual SLAM approaches are sensitive to illumination, with their precision drastically falling in dark conditions due to feature extractor limitations. The algorithms currently used to overcome this issue are not able to provide reliable results due to poor performance and noisiness, and the localization quality in dark conditions is still insufficient for practical use. In this paper, we present a novel SLAM method capable of working in low light using Generative Adversarial Network (GAN) preprocessing module to enhance the light conditions on input images, thus improving the localization robustness. The proposed algorithm was evaluated on a custom indoor dataset consisting of 14 sequences with varying illumination levels and ground truth data collected using a motion capture system. According to the experimental results, the reliability of the proposed approach remains high even in extremely low light conditions, providing 25.1\% tracking time on darkest sequences, whereas existing approaches achieve tracking only 0.6\% of the sequence time. 
\end{abstract}


\section{Introduction}
\subsection{Motivation}
Simultaneous localization and mapping (SLAM) algorithms are an inseparable part of every autonomous system. In visual-based approaches, the localization and mapping are performed using only the visual sources of information, without LiDARs and other active sensors. Visual methods of camera pose estimation have been actively researched in recent years and achieved robust and sustainable operation in academic and commercial use \cite{ondruvs2020autonomous}, \cite{jennings2020study}.

However, such methods strongly rely on the quality of input images: the localization precision significantly decreases in low light. This shortcoming is one of the main obstacles to visual SLAM algorithms application into outdoor mobile robotics: the outdoor light conditions are highly unpredictable, and, therefore, the reliability of such algorithms is insufficient for outdoor operation. The use of visual algorithms is challenging even in indoor conditions: for example, robots operating in warehouses \cite{kalinov2019high, kalinov2020warevision}, shopping rooms \cite{petrovsky2020customer} and hospitals \cite{perminov2021ultrabot, mikhailovskiy2021ultrabot} require a sufficient level of lighting, that, in case of fully unmanned facilities, increases the operation costs.

\begin{figure}[!t]
\centering
\begin{subfigure}[b]{0.24\textwidth}
\centering
\includegraphics[width=\textwidth]{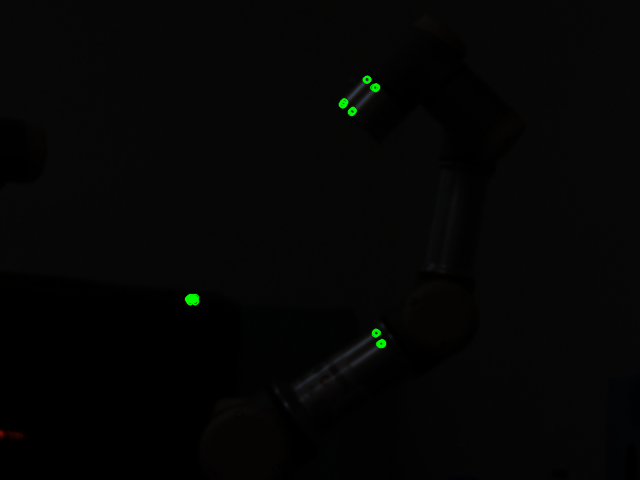} 
\caption{Original} \label{fig:sub1}
\vspace{1em}
\end{subfigure}
\begin{subfigure}[b]{0.24\textwidth}
\centering
\includegraphics[width=\textwidth]{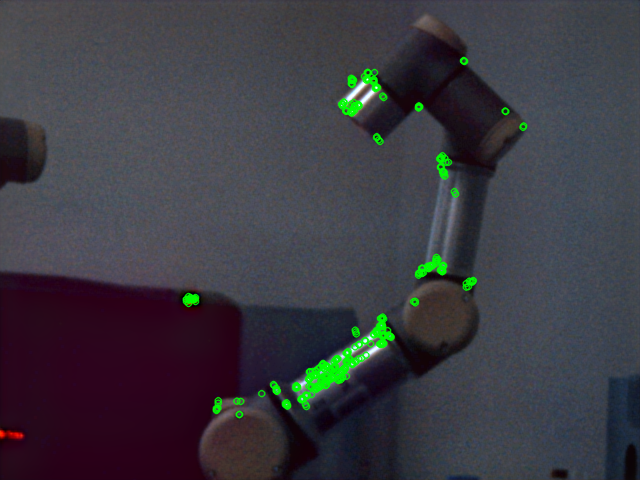} 
\caption{EnlightenGAN} \label{fig:sub2}
\vspace{1em}
\end{subfigure}
\begin{subfigure}[b]{0.24\textwidth}
\centering
\includegraphics[width=\textwidth]{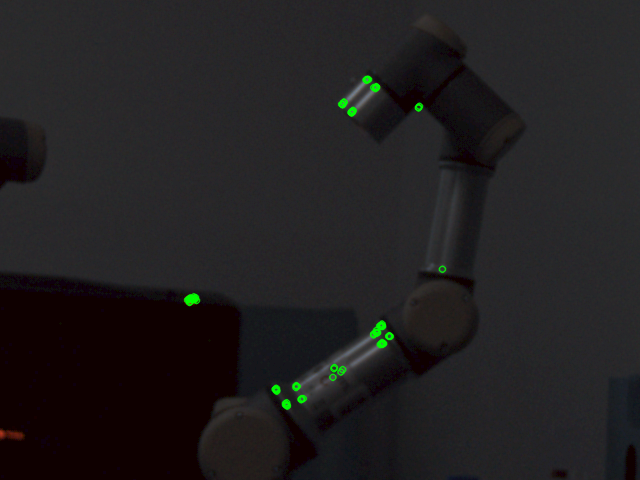} 
\caption{Gamma Correction, 2.0} \label{fig:sub3}
\end{subfigure}
\begin{subfigure}[b]{0.24\textwidth}
\centering
\includegraphics[width=\textwidth]{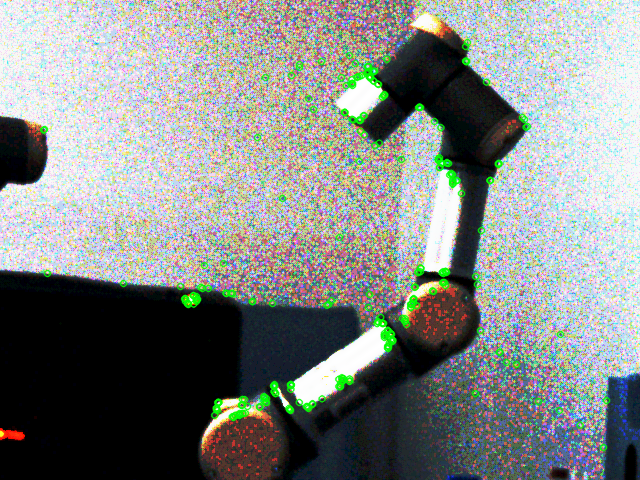} 
\caption{Histogram Equalization} \label{fig:sub4}
\end{subfigure}
\caption{Performance of DarkSLAM extractor on: (a) original image of UR3 robot, (b) image after proposed preprocessing, (c) image after gamma correction, (d) image after histogram equalization.}

\label{fig:preprocessing}
\end{figure}

\subsection{Problem Statement}
Visual SLAM approaches for autonomous systems must be invariant to environment light changes depicted on the input images. That includes night and day conditions, the presence of other cars' headlights, light from lampposts, traffic lights, reflections from traffic signs, and etc. The dataset collected by Pavlov et al. \cite{pavlov2019icevisionset} in Russian winter conditions clearly demonstrates what challenges in visual data processing must be coped with. Recent papers on autonomous outdoor robots show an increased interest in using only visible imaging sensors for navigation. Protasov et al. \cite{protasov2021cnn} developed a mobile robot with array of rolling-shutter cameras to achieve 360 deg. of field of view. Another approaches to dealing with the images in night conditions were proposed in \cite{karpyshev2021autonomous},  \cite{kalinov2021impedance}. However, the researchers either use the multispectral imaging or the additional IR infrastructure, which leads to a significant increase in the cost of an autonomous platform.

The problem of working with images taken in suboptimal conditions was addressed in multiple different approaches, most of which are focused on the input image preprocessing. The algorithms vary from simple mathematical transformations \cite{huang2012efficient}, \cite{pizer1987adaptive} to more complex and advanced methods \cite{dusmanu2019d2}, \cite{revaud2019r2d2}. These approaches have proven to slightly increase the flexibility of visual algorithms in dim conditions, however, they are still not able to achieve reliable results in extremely dark environments.

\subsection{Related Works}
Visual-based SLAM algorithms can be divided into two major groups. Direct, or feature-based, approaches \cite{ferrera2021ov} utilize the descriptors representing an image as a set of features. Indirect, or image-based, approaches \cite{min2021voldor}, in turn, process images as a whole instead of extracting features, for example, methods based on optical flow.

Each method has its own disadvantages: feature-based solutions are sensitive to the variance of the illumination level considered by the descriptor, whereas the image-based approaches fail for a quick change of the angle of view, which leads to lower accuracy of these methods. Most of the feature-based SLAM approaches such as \cite{ferrera2021ov}, \cite{mur2017orb}, \cite{li2020dxslam}, \cite{zhao2020good} have front-ends based on key-point feature extractors. While these methods work robustly in well-illuminated environments, standard descriptors lose their accuracy on low-light images. As a result, this leads to a decrease in localization accuracy. Thus, feature-based approaches are limited by the descriptor properties they utilize. The study of illumination variance on standard descriptors is carried out in several works \cite{ross2013novel}, \cite{ross2014method}, and it is shown that U-SIFT has the highest illumination invariance among standard feature extractors. However, these works study a limited number of extractors and do not include the efficient ORB extractor \cite{rublee2011orb} which is widely used in SLAM pipelines.

At the moment, there are learning-based methods of feature description, such as R2D2 \cite{revaud2019r2d2}, SuperPoint \cite{detone2018superpoint}, SuperGlue \cite{sarlin2020superglue}, nevertheless, they are not embedded inside the existing and proved SLAM algorithms due to the computational inefficacy of such algorithms.

Several works mitigate the issue of appearance change. For example, Sarlin et al. \cite{sarlin2021back} demonstrated the method capable of determining whether the feature is robust. The algorithm is able to find destructive features that are usually the regions of short-term entities such as fallen leaves. Thus, proposed approach is partially robust to season changes, although it can not be directly implemented to the low light cases. Another work \cite{hao2019lmvi} exploits image preprocessing with adaptive gamma correction and contrast adaptive histogram equalization to overcome the low-light conditions. However, the research lacks experimental results in dark conditions and was tested only on a dataset collected in a well-lit environment.

Closer to our work, \cite{porav2018adversarial} demonstrates the application of generative adversarial neural networks (GAN) preprocessing for improvement of localization metrics; they generate images from input sequence to make the weather conditions on generated images similar to the ones used for mapping. It proved that the localization in the map on generated images is more accurate than on real images. However, this work concentrates only on weather conditions and, therefore, the GAN has to be retrained for the task of dark image enhancement, which is almost impossible due to the lack of paired datasets for this purpose.

\subsection{Contribution}
In this paper, we propose a novel visual SLAM algorithm, dubbed DarkSLAM, capable of working robustly in low-light conditions without changing the classic ORB-SLAM2 \cite{mur2017orb} localization pipeline. This is achieved by utilizing the EnlightenGAN \cite{jiang2021enlightengan} image preprocessing module embedded into ORB-SLAM2 algorithm before the feature extraction procedure. Trained on an unpaired dataset, this generative network is able to enhance dark images, which eliminates the illumination limitation of the original method.

To evaluate the performance of the proposed approach, we collected an indoor dataset in varying illumination conditions. This dataset is unique in the number of dark conditions presented and ground truth data availability. The number of per-frame matches is evaluated on the unedited images from the dataset, and images preprocessed using the three enhancement approaches that have been previously proposed: histogram equalization, gamma correction, and the proposed GAN-based approach. Subsequently, the proposed algorithm is evaluated in terms of SLAM accuracy and convergence time. DarkSLAM is compared to ORB-SLAM2 launch on unedited data, and images preprocessed using the aforementioned methods. An example of the preprocessed images of UR3 collaborative robot is shown in Fig. \ref{fig:preprocessing}.

\section{Methodology}
\begin{figure*}[htbp]
\centerline{\includegraphics[scale=0.3]{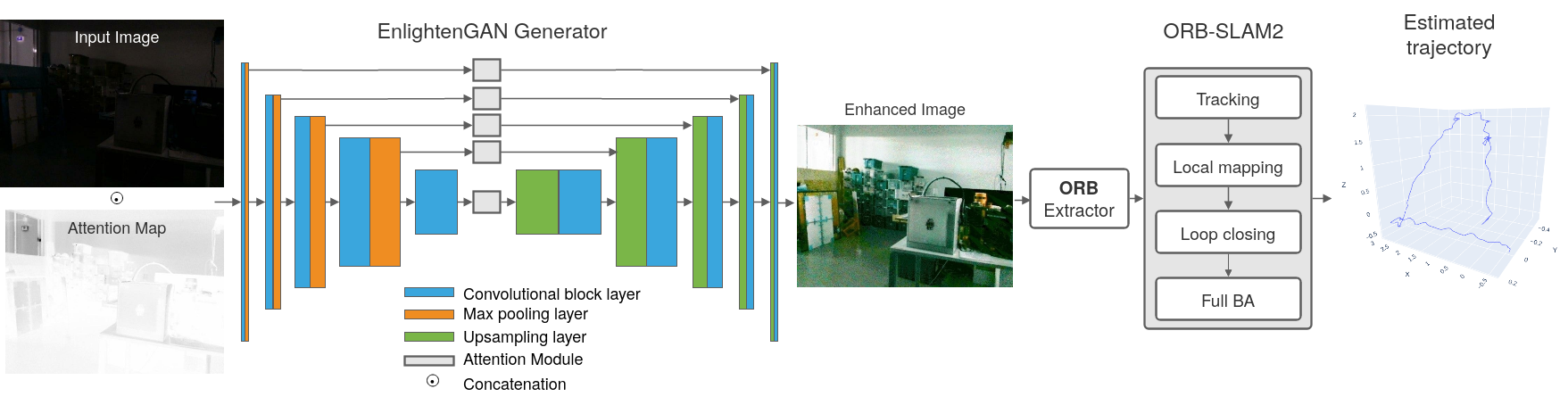}}

\caption{Structure of the proposed DarkSLAM approach.}
\label{fig:structure}
\end{figure*}

The proposed DarkSLAM pipeline is presented in Fig. \ref{fig:structure}. The raw images from the camera and corresponding illumination maps are transferred to the preprocessing module of the proposed algorithm, particularly to EnlightenGAN Generator; it outputs the images with significantly improved lighting. Then, the enhanced images are fed into the ORB feature detection algorithm \cite{rublee2011orb}, and, finally, the detected features are processed by ORB-SLAM2 localization pipeline, which estimates the camera positions, its trajectory and a sparse point cloud of the area surrounding the camera. 

\subsection{ORB-SLAM2}
ORB-SLAM2 \cite{mur2017orb} is a real-time SLAM algorithm for monocular, stereo, and RGB-D cameras, that reconstructs the camera trajectory and a sparse 3D point cloud of the camera field of view. It uses ORB feature extractor and descriptor \cite{rublee2011orb}, and utilizes loop closure based on Bag of Words \cite{galvez2012bags}. 

The system has a multithreaded structure. The first thread tracks the camera position with every frame by finding feature matches between the current image and the local map, and minimizing the reprojection error by preforming motion-only bundle adjustment (BA). The local mapping thread creates, edits and optimizes the local map using local bundle adjustment. The loop closure thread allows detecting large loops and correcting the accumulated drift through pose-graph optimization. This thread also performs full bundle adjustment after the pose-graph optimization, that allows to compute the optimal structure and motion solution.

Since this approach is feature-based, it requires multiple key points for initialization and precise tracking. In low-light conditions, the number of detected features decreases below the necessary amount, and the tracking algorithm fails to initialize. It happens due to ORB extractor limited variance to illuminations. The proposed approach mitigates this problem by introducing an embedded GAN preprocessing module that improves the light on input images and results in better initialization and tracking in dark and dim environments.

\subsection{EnlightenGAN}
EnlightenGAN \cite{jiang2021enlightengan} is the generative adversarial network proposed to enhance low-light images with heterogeneous illumination levels.
It uses the U-Net architecture as a generator network. The U-Net based generator is implemented with eight convolutional layers. Each block is 3×3, followed by LeakyReLu and batch normalization layers. One of the upsampling deconvolutional layers is replaced with one bilinear upsampling layer and followed convolutional one for mitigation of checkerboard artifacts that appears in classical methods such as a histogram normalization \cite{pizer1987adaptive}.

Adding the attention map, which is the inverse of the grayscale image, to the input and element-wisely multiplying and adding it to the generator's output helps to define the regions of low and high light levels, enhancing them as self-regularization. The structure of the local discriminator network is inspired by PatchGAN \cite{yu2018generative}. It utilizes the decomposition of real and fake images as patches that allows avoiding severe color distortion and uneven illumination. 

The global discriminator is defined as the relativistic discriminator structure \cite{jolicoeur2018relativistic} that allows to distinguish more realistic images between real and generated, and guides the generator network to create the realistic images. The discriminator's standard loss function is reformulated as the least square loss proposed in LSGAN \cite{mao2017least}. The EnlightenGAN also implements the loss between an input image and enhanced one \cite{johnson2016perceptual} that allows to constraint the extracted VGG-feature distance and to keep the perceptual similarity of two pictures. We use a pretrained model of EnlightenGAN for image enhancement. However, our method is appropriate for fine-tuning the GAN network in a specific environment.

\section{Experiments and Results}

\subsection{Experimental Setup}

\begin{figure}[ht]
\centering
\includegraphics[width=0.25\textwidth]{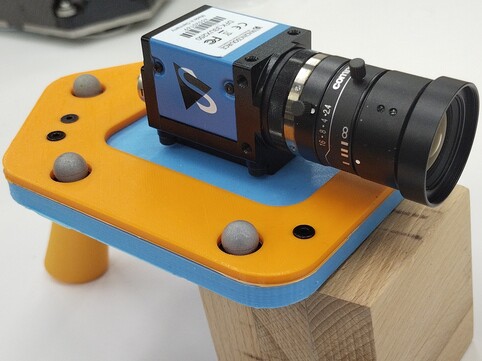}
\caption{The Imaging Source DFK 33UX250 global-shutter camera with six spherical motion-capture markers.}
\label{fig:holder}
\end{figure}

To evaluate the proposed DarkSLAM approach, we have collected th dataset containing image sequences in various light conditions. The visual data was recorded using the Imaging Source DFK33UX250 global-shutter camera. The ground truth data for camera position was collected using the Vicon Vantage V5 motion capture system. This system consists of 12 fixed IR cameras for multi-view position calculation and is capable of sub-millimeter precision at 100 fps. 
We have designed a rig containing a camera and six reflective markers recognized by the mocap system. The marker pattern is asymmetrical, ensuring that not only the position of the rig, but also its orientation, are correctly recorded. Image of the rig is presented in Fig. \ref{fig:holder}.

To evaluate the precision of DarkSLAM, the transformation between the Vicon mocap system coordinates and camera coordinates needs to be calculated. The transformation between the camera rig and the Vicon system was provided by the Vicon software. The transformation between the camera rig and camera itself was calculated using the calibration method based on chessboard tile detection. A flat surface with a chessboard pattern was supplied with reflective Vicon markers, and, by knowing the position of the rig and the chessboard in Vicon coordinates, and the position of the chessboard in the camera coordinates, we obtained the transformation matrix between the Vicon and the camera using the PnP method \cite{lepetit2009epnp}. It allowed us to compare the estimated trajectories with the ground truth data. Moreover, application of monocular camera in the setup requires the SLAM-estimated trajectories to be aligned and scaled for precise evaluation of accuracy metrics. For this purpose, the Evo package \cite{evo} was applied to align, scale, and synchronize the timestamps of two trajectories.

\subsection{Dataset}\label{AA}

\begin{figure}[t]
    \centering
    \includegraphics[width=0.15\textwidth]{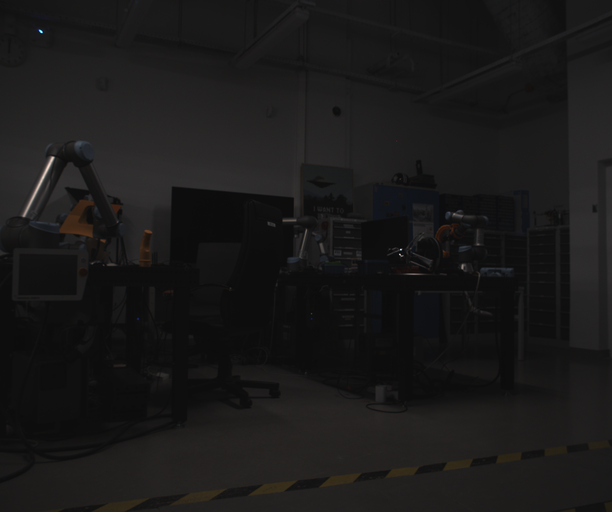}
    \centering
    \includegraphics[width=0.15\textwidth]{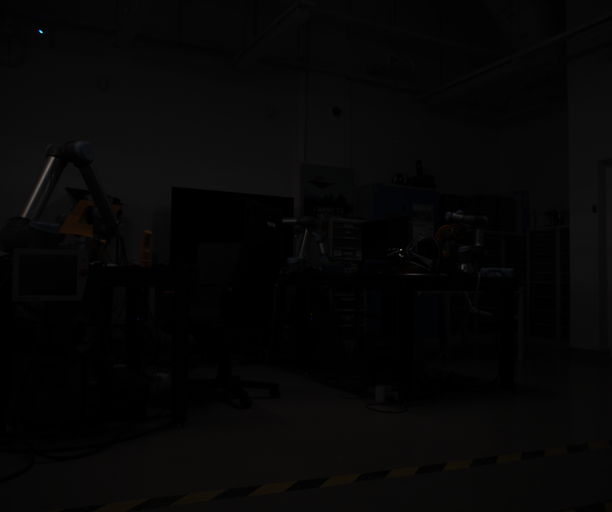}
    \centering
    \includegraphics[width=0.15\textwidth]{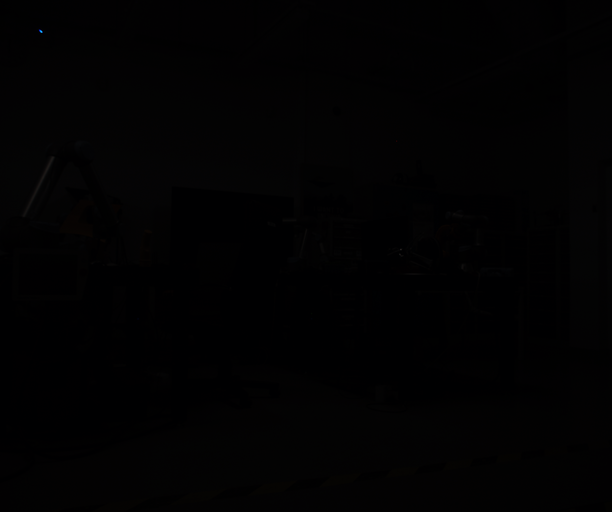}
    \caption{Examples of images in the Shaded (left), Semi-dark (center) and Dark conditions (right).}\label{fig:conditions}
\end{figure}

The collected dataset consists of 14 video sequences in various illumination conditions. Examples of illumination levels in the dataset are presented in Fig. \ref{fig:conditions}. The luminosity in the room was varied within the following ranges: for Shaded sequence: from 5 to 9 lx., for Semi-dark sequence: from 3 to 5 lx., for Dark sequence: from  0 to 2 lx.
 
Dataset images were downscaled to the resolution of 640×480 pixels, the recording was performed at 20 fps. Exposure and light balance were fixed, and the gain parameter was set to zero to ensure consistent experimental results. Before data collection, the camera was calibrated using the ROS camera calibration package \cite{camcal}.
 
All sequences were recorded with the same camera parameters in a room with controlled illumination, ensuring that the collected images represent various levels of luminosity. The camera was moved along a rectangular trajectory manually. The length of each sequence was approximately the same.

To validate the performance of DarkSLAM, we used the obtained datasets. Since our work is primarily aimed at investigating the improvement of localization in low-illuminated environments, it was divided into two parts. Firstly, we evaluated the ability of ORB extractor \cite{rublee2011orb} to find keypoints on the low-light dataset images. Secondly, we tested the proposed DarkSLAM approach according to standard SLAM metrics \cite{shi2020we}.

\subsection{Feature Detection Experiment}\label{AA}
The first experiment is devoted to validate whether the light change and preprocessing affect the ORB feature robustness. The experiment was carried out on the collected dataset sequences. Each sequence was processed using a set of image lightening methods: histogram equalization \cite{pizer1987adaptive}, gamma correction \cite{huang2012efficient} with gamma values of two and four, and EnlightenGAN. The aim of this experiment is to compare the number of feature matches between consecutive images processed with the aforementioned methods, and original ones.

ORB keypoint descriptors were calculated on these four types of processed images and original ones using the same settings as described in \cite{rublee2011orb}. To determine whether a feature is good, we evaluated the matched keypoint descriptors by compiling a fundamental matrix \cite{luong1996fundamental}. The average numbers of matched descriptors between frames for each of the light levels are listed in Table \ref{table:feature_metrics}.

\begin{figure*}[htbp]
\centerline{\includegraphics[scale=0.65]{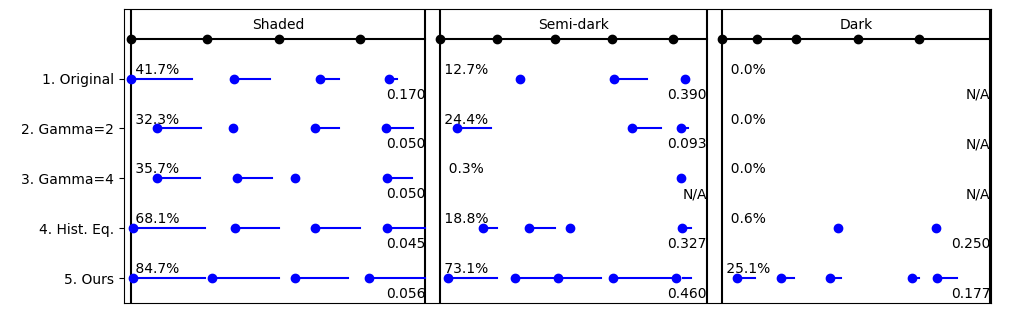}}
\caption{Per-sequence testing results. Black dots represent different sequences of one luminosity level. Blue dots represent SLAM initialization, blue lines represent the initialization length. The values on the right of each sequence group is average ATE RMSE metric, [m]; on the left – CR metric, \%.}
\label{fig:loris}
\end{figure*}

\begin{table}[t!]
\caption{Average number of feature matches between the consecutive frames.}
\begin{center}
\begin{tabular}{|p{1.5cm}|p{0.9cm}|p{0.9cm}|p{0.9cm}|p{0.9cm}|p{0.9cm}|}
\hline
\textbf{}&\multicolumn{5}{|c|}{\textbf{Processing methods}} \\
\cline{2-6} 
\textbf{Sequence} & \textbf{\textit{Original}} & \textbf{\textit{GAN}} &  \textbf{\textit{Hist.}} & \textbf{\textit{Gamma corr., $\gamma$=2}} & \textbf{\textit{Gamma corr., $\gamma$=4}} \\
\hline
  00 shaded & 178 & 292 & 282 & 205 & 166   \\
  01 shaded & 60 & 215 & 274 & 82 & 56   \\
  02 shaded & 94 & 158 & 265 & 105 & 91   \\
  03 shaded & 101 & 253 & 275 & 128 & 100   \\
\hline
  04 semi-dark & 114 & 262 & 272 & 144 & 108   \\
  05 semi-dark & 92 & 205 & 267 & 109 & 91   \\
  06 semi-dark & 59 & 208 & 272 & 80 & 59   \\
  07 semi-dark & 105 & 309 & 301 & 151 & 105   \\
  08 semi-dark & 83 & 173 & 253 & 94 & 72   \\
\hline
  09 dark & 91 & 177 & 258 & 102 & 90   \\
  10 dark & 76 & 165 & 255 & 87 & 75   \\
  11 dark & 37 & 133 & 247 & 52 & 40   \\
  12 dark & 21 & 101 & 254 & 28 & 24   \\
  13 dark & 34 & 161 & 254 & 49 & 36   \\
\hline
\end{tabular}
\label{table:feature_metrics}
\end{center}
\end{table}

The ``Original" column in Table \ref{table:feature_metrics} demonstrates that robustness of ORB features degrades with the conditions getting darker; it constitutes 81 matches on average. Regarding the preprocessing techniques, images processed using histogram equalization and EnlightenGAN have 200 and 266 matched keypoints on average, respectively, which is two times greater than for the original dark images and gamma correction approaches (101 and 82 matched keypoints,  respectively). Thus, the GAN preprocessing enables to improve the decrease of ORB feature extractor robustness in dark conditions.

\subsection{SLAM Performance Experiment}\label{AA}
The second experiment was carried out to validate whether the use of preprocessing improves the localization accuracy of ORB-SLAM2 method.
To evaluate the accuracy of the proposed method, the Absolute Trajectory Error (ATE) and Correct Rate (CR) \cite{shi2020we} metrics were used. CR demonstrates the operating time of the SLAM algorithm at which the values of ATE and Absolute Orientation Error (AOE) do not exceed the specified thresholds, meaning the complete loss of tracking. Larger CR metric means longer performance of the SLAM algorithm, and smaller ATE metric means more accurate localization. However, these two metrics do not correlate: the ATE is likely to increase gradually with the time. Thereby, the increase in ATE metric may be caused by the much longer convergence of the operation time.

Fig. \ref{fig:loris} shows the results obtained during the experiment. The original ORB-SLAM2 can initialize and shortly work only for the brightest sequences of the dataset, denoted as Shaded. Its CR metric drops with the decrease of sequence luminosity from 41 to 0\% from Shaded to Dark sequences. Thus, its performance degrades with the sequence conditions getting darker, the same as in the Feature Detection Experiment. Likewise, gamma correction method with factors of two and four has only a slight improvement in Semi-dark sequence compared to the original ORB-SLAM2. The histogram equalization approach refines both CR and ATE metrics relative to the original method; on the Semi-dark sequences, it was able to increase CR from 12.7\% to 18.8\% while decreasing the ATE by 6.3 cm. However, this approach is comparable to the proposed one only in the Shaded sequence, while on Dark sequence its metrics drop drastically and constitute only 25 cm and 0.6\% for accuracy and operation time, respectively.

Dark-SLAM approach shows the best tracking time and CR metric regardless of the luminosity levels. It can track at least 25.1\% of trajectory even for the hardest Dark sequence, while its ATE metric stays accurate and constitutes 17.7 cm.

\section{Conclusions and Discussion}

We proposed a novel approach to increase the robustness of visual SLAM in dark conditions with the help of the embedded preprocessing module based on a generative adversarial network. The proposed pipeline was evaluated on the dataset collected in different levels of illumination, and the results were compared with techniques widely used for low-light image enhancement. 
The experiment on the number of matches for ORB features has shown the efficiency  of the proposed preprocessing module; although histogram equalization method has demonstrated the high number of matches as well, the quality of these features is unreliable for localization purposes due to the presence of checkerboard artifacts and an increased amount of noise on the processed images.
Experimental results of DarkSLAM revealed that, despite the high number of matches for histogram equalization method in the feature matching experiment, it is more robust and outperforms all aforementioned approaches with the operation time (CR metric) by 16.6\% in Shaded sequences, by 54.3\% in Semi-dark sequences and by 24.5\% in Dark sequences, in which the luminosity varied from 0 to 2 lx. The most significant improvement was found in Dark sequences, where the second-best performing algorithm, histogram equalization, was only able to sustain SLAM convergence for 0.6\% of the time, and the other approaches did not converge at all. Despite the use of the ATE RMSE metric in the original paper \cite{shi2020we}, we consider this metric to be non-representative when the time of correct SLAM operation, and the trajectory length, noticeably varies between different approaches.



\bibliographystyle{IEEEtran}
\bibliography{literature.bib}

\end{document}